\documentclass[11pt]{article}

\usepackage{graphicx}
\usepackage{fullpage}
\usepackage{authblk}

\usepackage{float}
\usepackage{subcaption}

\usepackage{listings}
\usepackage{xcolor}

\definecolor{vscodebg}{RGB}{248,248,248}
\definecolor{vscodegray}{RGB}{110,110,110}

\usepackage[pagebackref=false,breaklinks=true,colorlinks,urlcolor=blue,citecolor=blue,linkcolor=blue,bookmarks=false]{hyperref}
\usepackage[numbers,sort&compress]{natbib}

\title{NIMO Controller: a self-driving laboratory orchestrator\\based on the Model Context Protocol}

\author[1]{Naruki Yoshikawa\thanks{Correspondence: \texttt{yoshikawa.naruki@nims.go.jp}}}
\author[1,2]{Ryo Tamura}

\affil[1]{National Institute for Materials Science, Tsukuba, Ibaraki, Japan}
\affil[2]{Graduate School of Frontier Sciences, The University of Tokyo, Chiba, Japan}
\date{}

\begin{document}

\maketitle

\begin{abstract}
Self-driving laboratories (SDLs) are attracting increasing attention as a means of accelerating scientific discovery; however, developing SDL software remains technically demanding. To improve accessibility, orchestration software frameworks have been proposed to coordinate SDL components, but many existing frameworks are primarily designed for human interaction and lack standardized interfaces for direct integration with AI agents. In this work, we propose an SDL software architecture based on the Model Context Protocol (MCP), in which all SDL functionalities are exposed through MCP servers. Following this design principle, we introduce NIMO Controller, an MCP-based SDL orchestrator that integrates the experimental planning software NIMO. It provides a visual programming interface automatically generated through MCP-based tool discovery, allowing human users to design experimental workflows without writing code. The same MCP backend is also accessible to AI agents, providing a unified interface through which both human users and AI agents can interact with SDL components. AI agents can invoke MCP tools directly or generate NIMO workflows from natural language instructions. We evaluated the agentic workflow generation performance of various LLMs and conducted a user study using a color-matching SDL.
\end{abstract}

\section{Introduction}
Self-driving laboratories (SDLs) combine automated experimentation with data-driven experimental design to accelerate scientific discovery~\cite{tom2024self}.
Despite growing interest, the practical deployment of SDLs still faces several challenges.
Among these, software development remains a major barrier to accessibility: most SDL software requires programming expertise, limiting its adoption by domain experts who design experiments but do not necessarily have programming skills.
To address this issue, various orchestration software frameworks have been proposed to facilitate the development of SDL software~\cite{roch2020chemos,malcolm2024chemos,fei2024alabos}.
We have previously reported an integration of IvoryOS~\cite{zhang2025ivoryos} and NIMO~\cite{tamura2026nimo} (formerly NIMS-OS~\cite{tamura2023nims}) that provides a graphical user interface for laboratory workflow design that incorporates algorithmic decision-making~\cite{yoshikawa2026bridging}.
Although IvoryOS automatically generates an accessible interface for human scientists from Python code, SDLs often need to integrate legacy devices that may not support Python~\cite{tamura2025seamless}. Furthermore, with the growing interest in AI scientists, accessibility of SDL functionalities to AI systems should also be taken into consideration.

To enable interoperability between large language models (LLMs) and external tools, the Model Context Protocol (MCP) has recently been proposed~\cite{mcp_spec}. MCP standardizes the interface through which external tools and data sources are exposed to LLMs, making it a promising foundation for an SDL abstraction layer accessible to AI systems. In this paper, we propose an MCP-centric SDL software architecture, in which both laboratory hardware and decision-making algorithms are exposed as MCP tools. Based on this design, we introduce a new SDL orchestrator, NIMO Controller. It lowers the entry barrier to NIMO's experimental planning algorithms through a visual programming interface automatically generated from MCP servers. The embedded AI agent assists users by operating the SDL and retrieving information through MCP tool calls, designing workflows in the visual programming interface, and generating summary reports. The key contributions of this paper are as follows:
\begin{itemize}
\item \textbf{MCP-based abstraction of SDL components}:
NIMO Controller adopts MCP to encapsulate SDL components, making them accessible to AI agents with minimal integration effort. This abstraction layer provides loose coupling between the orchestration software and SDL components, enabling independent and language-agnostic development of individual components. New devices and algorithms can be added simply by launching additional MCP servers without modifying the client code, and the same mechanism allows remote experiments to be conducted through remote MCP servers.

\item \textbf{Automatic generation of visual programming interface}: To make SDLs accessible to users without programming expertise, NIMO Controller automatically generates a block-based visual programming interface through MCP tool discovery. Users can design experimental workflows by composing draggable blocks without writing code. This lowers the entry barrier for domain experts while also facilitating the use of SDLs for educational purposes. These blocks are generated from the same MCP tools invoked by AI agents, allowing human users and AI agents to control experiments through a common backend interface.

\item \textbf{Agentic workflow design}: To enable AI-assisted creation of complex experimental planning workflows in NIMO, we introduce a structured workflow representation and an AI agent that generates NIMO workflows from natural language instructions. The generated workflow is converted into a Blockly representation, allowing users to inspect and edit it before execution. The executed workflow is subsequently serialized into the NIMO workflow XML format for reproducibility and reporting.
\end{itemize}

\section{Methods} \label{sec:methods}
In this section, we describe the MCP-based SDL software architecture, the structured representation of NIMO workflows, the implementation of NIMO Controller, and its agentic workflow design mechanism.
The description of MCP in this article is based on the MCP specification version 2025-11-25~\cite{mcp_spec}.

\subsection{MCP-based software architecture for SDLs}
An SDL is composed of multiple components, such as robots for automating experiments, software for experimental planning or simulation, and external data resources.
These components may run as separate processes on a single computer or on separate machines at remote locations.
Among the various communication protocols available for connecting these components, we adopt the Model Context Protocol (MCP) for the following reasons.
\begin{itemize}
    \item \textbf{Readiness for AI agents}: MCP provides a standardized interface for connecting external tools with LLMs. As a result, many AI agent frameworks support MCP-based tool calling, facilitating the integration of AI agents with SDLs.
    \item \textbf{Openness and portability}: The MCP specification is freely available, and SDKs are available for many programming languages across a wide range of operating systems. Because SDLs often rely on legacy systems that do not support Python~\cite{tamura2025seamless} or that run only on specific operating systems, this portability is beneficial for connecting heterogeneous components to a central orchestrator. In addition, MCP provides the Streamable HTTP transport, allowing MCP servers to be hosted on remote machines and thereby facilitating remote experiments.
\end{itemize}

\subsection{NIMO workflow}
Conducting an experiment requires organizing multiple unit operations into a workflow~\cite{gottstein2026foundational}.
A typical experiment using NIMO consists of the following steps~\cite{tamura2026nimo}:
\begin{enumerate}
    \item Users prepare a candidate CSV file, typically named \verb|candidates.csv|, that enumerates all candidate experimental conditions, and provide the file to NIMO.
    \item NIMO selects the next experimental condition with \verb|nimo.selection()| using the algorithm specified by the user.
    \item The automated experiment system performs the experiment based on the proposed experimental conditions.
    \item NIMO updates the candidate file with \verb|nimo.output_update()|, which receives the experimental results corresponding to the proposed conditions.
    \item Steps 2--4 are repeated for a specified number of iterations, and the results are visualized using NIMO's visualization functions.
\end{enumerate}

To systematize the description of this workflow, we developed an XML-based workflow description language for NIMO.
Listing~\ref{lst:nimo-workflow} shows an example of a NIMO workflow for a color-matching experiment that uses a Bayesian optimization algorithm to optimize the mixing ratio of three primary colors for reproducing a target color.
The XML is composed of three sections:
\begin{itemize}
    \item \verb|<servers>|: This section describes the MCP servers used in the experiment, and lists the names and parameters of the available tools provided by each server. Optionally, the versions of the MCP servers can be recorded.
    This section specifies the expected MCP server interfaces and is used to validate the workflow description.
    \item \verb|<candidates>|: This section lists the parameter and objective names used to validate the workflow description. Optionally, the file name of the candidate CSV file and its SHA256 hash at the beginning of the experiment can be recorded. 
    \item \verb|<workflow>|: This section describes the workflow that calls NIMO tools and MCP servers. NIMO tools are invoked through a dedicated \verb|<nimo>| tag, indicating that NIMO functionality is used. MCP server functionality is invoked through the \verb|<call>| tag.
    Note that the \verb|<nimo tool="update">| tag contains a single MCP tool call whose result is used to update the candidate CSV file. The called MCP tool must return a single floating-point or integer value.
\end{itemize}

\begin{lstlisting}[
  language=XML,
  caption={Example of a NIMO workflow for a color-matching experiment described in XML. This example is taken from the benchmark tasks in Section~\ref{sec:workflow} and differs from the workflow actually used in the case study in Section~\ref{sec:color_matching}.},
  label={lst:nimo-workflow},
  float=ht
]
<nimo-workflow version="0.1.0">
  <servers>
    <server name="color">
      <tool name="move_to_home"/>
      <tool name="move_to_color_well" params="{color: str}"/>
      <tool name="move_to_mix_well" params="{well: int}"/>
      <tool name="aspirate" params="{volume: float}"/>
      <tool name="dispense" params="{volume: float}"/>
      <tool name="get_color_diff" params="{well: int, target_hex: str}"/>
    </server>
  </servers>
  <candidates>
    <parameter name="blue"/>
    <parameter name="yellow"/>
    <parameter name="red"/>
    <objective name="objective"/>
  </candidates>
  <workflow>
    <call server="color" tool="move_to_home"/>
    <repeat counter="i" times="12">
      <if counter="i" op="&lt;" value="4">
        <then><nimo tool="selection" method="RE"/></then>
        <else><nimo tool="selection" method="PHYSBO" minimization="true"/></else>
      </if>
      <call server="color" tool="move_to_color_well" color="red"/>
      <call server="color" tool="aspirate" volume="proposal.red"/>
      <call server="color" tool="move_to_mix_well" well="counter.i"/>
      <call server="color" tool="dispense" volume="proposal.red"/>
      <call server="color" tool="move_to_color_well" color="blue"/>
      <call server="color" tool="aspirate" volume="proposal.blue"/>
      <call server="color" tool="move_to_mix_well" well="counter.i"/>
      <call server="color" tool="dispense" volume="proposal.blue"/>
      <call server="color" tool="move_to_color_well" color="yellow"/>
      <call server="color" tool="aspirate" volume="proposal.yellow"/>
      <call server="color" tool="move_to_mix_well" well="counter.i"/>
      <call server="color" tool="dispense" volume="proposal.yellow"/>
      <nimo tool="update">
        <call server="color" tool="get_color_diff" well="counter.i" target_hex="00ffff"/>
      </nimo>
    </repeat>
    <nimo tool="plot_history_best" minimization="true"/>
  </workflow>
</nimo-workflow>
\end{lstlisting}

This design is inspired by the Chemical Description Language (XDL)~\cite{mehr2020universal}.
Unlike XDL, which defines a fixed set of steps, this language allows the set of invocable operations to be easily extended by adding MCP servers.
The implementation of each MCP server is left to the developer, so different types of hardware can be used as long as they provide the same tool interface.
This design improves portability while also contributing to reproducibility by reducing reliance on ambiguous natural language descriptions of experimental procedures.

The expressiveness of the NIMO workflow description language is intentionally limited to predefined workflow structures and MCP tool calls. This restriction ensures that agent-generated workflows can only perform operations provided by the workflow language and the available MCP tools. Consequently, their execution can be controlled at the level of individual MCP tool calls, for example, through user approval or safety mechanisms implemented in each tool.

\subsection{NIMO Controller}
\begin{figure}[ht]
\centering
\includegraphics[width=0.95\textwidth]{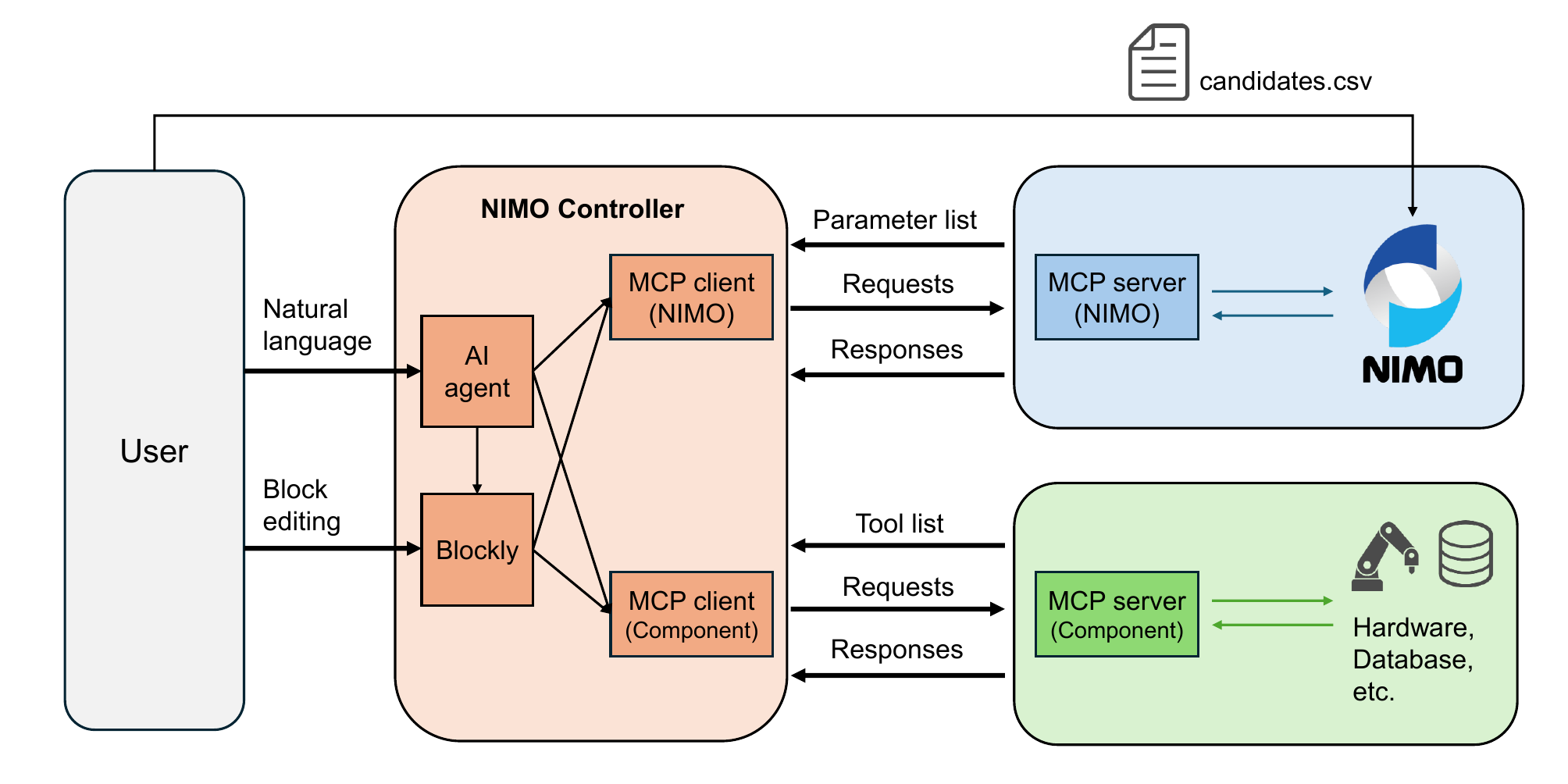}
\caption{Architecture of NIMO Controller. NIMO Controller serves as an MCP host that integrates a natural language interface (via an AI agent) and a visual block-based interface (built on Blockly) into a unified frontend. Through MCP clients, these interfaces interact with backend MCP servers: the NIMO MCP server exposing decision-making algorithms, and component MCP servers exposing laboratory hardware and other SDL resources.}
\label{fig:overview}
\end{figure}

In this section, we describe the design of NIMO Controller and its embedded AI agent.
NIMO Controller provides a unified frontend that integrates a visual programming interface and a natural language interface in a single application.
On the backend, NIMO Controller communicates with MCP servers to support experimental planning and automated experimentation.
The architecture of NIMO Controller is shown in Figure~\ref{fig:overview}.

Users can manually design a NIMO workflow in the Blockly interface by dragging and dropping blocks, while the embedded AI agent can automatically invoke MCP tools through the natural language interface. The agent can also generate a structured NIMO workflow, which is rendered in the Blockly interface for user inspection before execution. Users can then monitor its execution status through the same interface.

\subsubsection{NIMO MCP server}
NIMO Controller includes a built-in MCP server for NIMO~\cite{tamura2026nimo} as its experimental planning engine.
NIMO tools are exposed through this MCP server using the stdio transport to minimize communication overhead.
The user first uploads a CSV file (\texttt{candidates.csv}) that lists the candidate experimental conditions through the user interface.
The parameter names and their current values are shown to both the user and the agent.
Other key functions, such as the \texttt{selection()} tool, which proposes the next experimental conditions, and the \texttt{update()} tool, which updates NIMO's internal state with the latest experimental results, are provided.
Visualization tools are also exposed; they return graphical representations of experimental results as images, which are displayed in the user interface.

\subsubsection{Component MCP servers}
NIMO Controller can be connected to any number of component MCP servers, each of which encapsulates specific functionality required by the experimental workflow, such as laboratory robots or database services. NIMO Controller communicates with these servers using the Streamable HTTP transport, allowing them to run on either local or remote machines. These servers are typically implemented by wrapping existing APIs for laboratory instruments. Since the implementation of each MCP server is left to the developers, they can be developed independently of the frontend.

At the time of writing, the component MCP servers supported by NIMO Controller have the following constraints. Among the three primitives defined by MCP (tools, resources, and prompts), only tools are supported. The output of MCP tools should be of a type that can be displayed as text or images.
For the visual programming interface, the input schema is limited to basic data types supported by Blockly, such as numbers and strings.
MCP tools used to provide experimental results to NIMO's \texttt{update()} tool, which updates the candidate CSV file, must return a single floating-point or integer value.
Because authentication is not currently implemented, we recommend restricting access to MCP servers by using a secure private network or an encrypted channel, such as a VPN or an SSH tunnel.

\subsubsection{Visual programming interface}
NIMO Controller provides a visual programming interface based on Blockly, where users can construct workflows by dragging and dropping blocks.
The interface provides several types of toolboxes: the control toolbox contains blocks for controlling the experimental workflow, such as repeat and conditional control blocks; the NIMO toolbox provides blocks for accessing NIMO's decision-making and visualization functionality; and each connected MCP server has a dedicated toolbox containing blocks for the SDL functions exposed by that server.

The blocks for MCP server functions are automatically generated from the tool definitions retrieved from the corresponding MCP servers.
When the page is loaded, the frontend retrieves the tool list from each MCP server registered during the MCP initialization phase.
For each discovered tool, a Blockly block definition is dynamically generated, with input fields such as numeric or text fields derived from the tool's input schema.
This mechanism allows the visual programming interface to automatically reflect the tools provided by connected MCP servers without requiring manual modification of the frontend.

When the user clicks the \texttt{Run} button, the designed workflow is sent to the backend and executed through calls to the corresponding MCP servers.
The interface displays the designed workflow in XML format, and it highlights the currently executing block in real time, allowing users to inspect the workflow and monitor its execution.

\subsubsection{Natural language interface}
In addition to the visual programming interface, NIMO Controller provides a natural language interface, in which users interact with an AI agent via a chat window. The agent is implemented using Pydantic AI~\cite{pydantic_ai} and is connected to the same NIMO MCP server and component MCP servers as the visual programming interface. Users can select the underlying LLM from the chat window, including cloud-based models provided by OpenAI and Anthropic as well as local models served by Ollama.

The agent can respond to user queries either directly using the LLM's own knowledge or by invoking tools. To ensure safe operation, users are asked to approve or reject each tool call requested by the agent. For autonomous operation, users can enable an auto-approve option that allows tool calls to proceed without user intervention. When a user's request requires a sequence of tool calls, the agent can instead generate and execute a workflow. 
This workflow generation mechanism is described in Section~\ref{sec:agent}.

\begin{figure}[ht]
\centering
\includegraphics[width=0.95\linewidth]{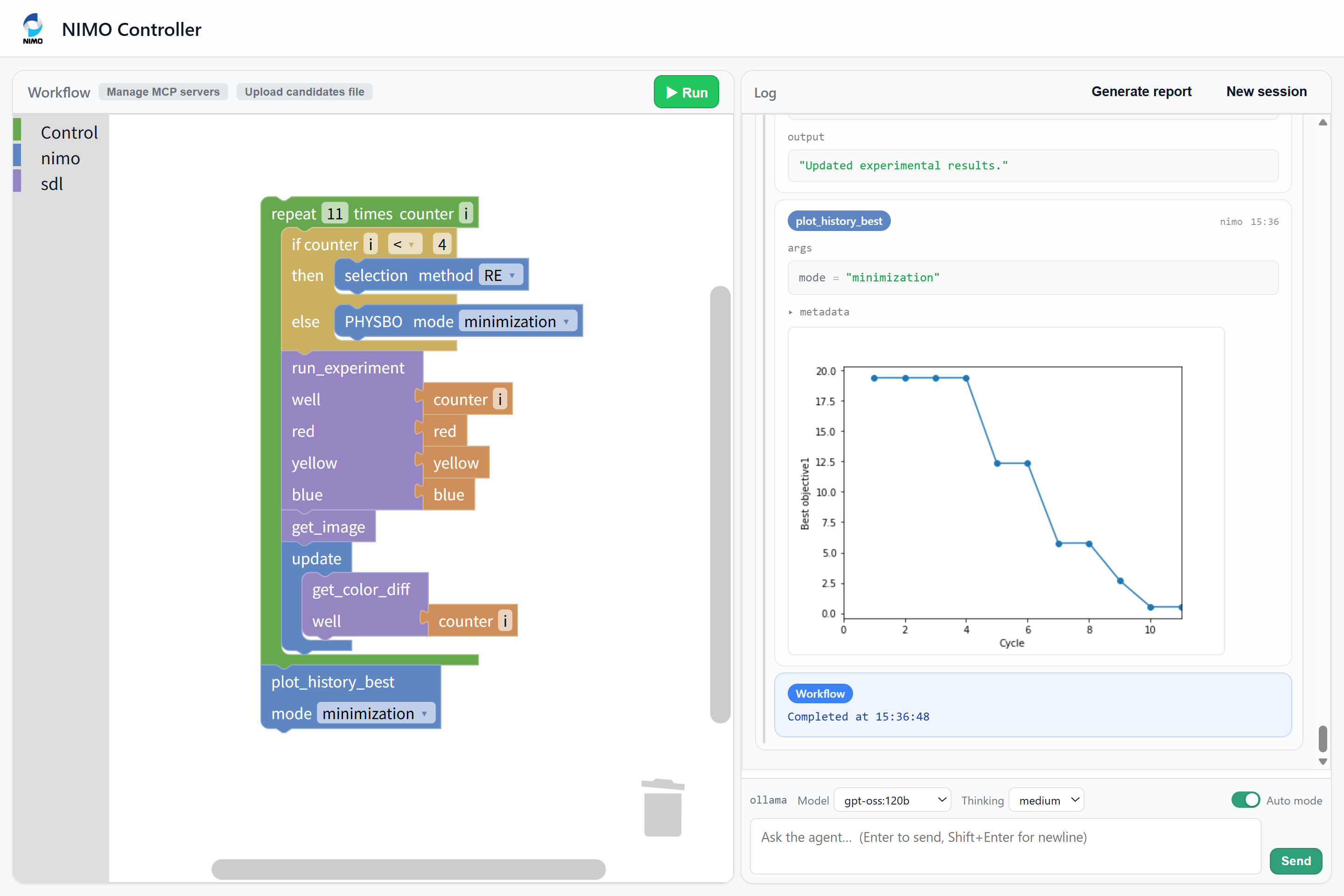}
    \caption{A screenshot of NIMO Controller. The left side shows the experimental workflow designed in Blockly, while the right side provides execution logs, including the output of each block. Users can interact with the AI agent through the chat box to invoke MCP tools or generate a workflow.}
\label{fig:screenshot}
\end{figure}

To provide both interfaces within a unified user interface, NIMO Controller uses a two-column layout (Figure~\ref{fig:screenshot}). The left column provides the Blockly interface, where users can manually construct and execute NIMO workflows. The right column provides execution logs and the natural language interface, allowing users to interact with the AI agent, invoke MCP tools, request workflow generation, and monitor execution status.

\subsubsection{Session report}
NIMO Controller organizes experimental workflows into sessions, each of which can contain multiple workflow runs and individual MCP tool calls.
A new session is created when a user uploads a new candidate CSV file or explicitly starts a new session.
When a session is created, NIMO Controller sets up a dedicated directory for the session, where all intermediate files and execution logs generated during the session are stored.
The uploaded candidate CSV file is copied to this directory, and NIMO tools read from and update this copy throughout the session.

NIMO Controller provides an automated report generation function for each session.
When the user requests a report, the LLM analyzes the experimental history recorded in the session, including the execution logs and associated experimental data, and generates a summary of the session.
The generated report is provided in both HTML and Markdown formats, allowing users to review and edit the report and convert it to other formats as needed.

\subsection{Agentic workflow design}\label{sec:agent}
Although MCP provides a unified interface for tool calling, repeatedly invoking MCP tools through an LLM can incur substantial token usage and computational overhead.
When an LLM invokes a tool through MCP, tokens are consumed both to construct the tool call and to process the returned result.
For workflows involving many sequential or repetitive operations, this overhead can increase both the cost and execution time of the experiment.
Moreover, the workflow planned by the agent is not visible to users and therefore cannot be inspected before execution.

To address these limitations, we introduce an agentic workflow design mechanism inspired by the concept of code execution with MCP~\cite{code_execution_mcp}. Instead of invoking MCP tools individually through the LLM during execution, the agent first designs a structured workflow, which is then executed by a deterministic workflow executor. Consequently, workflow execution requires no additional LLM inference or token consumption, making this approach more efficient for workflows involving multiple tool calls. Users can review and modify the generated workflow before execution.

\begin{figure}[ht]
\centering
\includegraphics[width=1.0\linewidth]{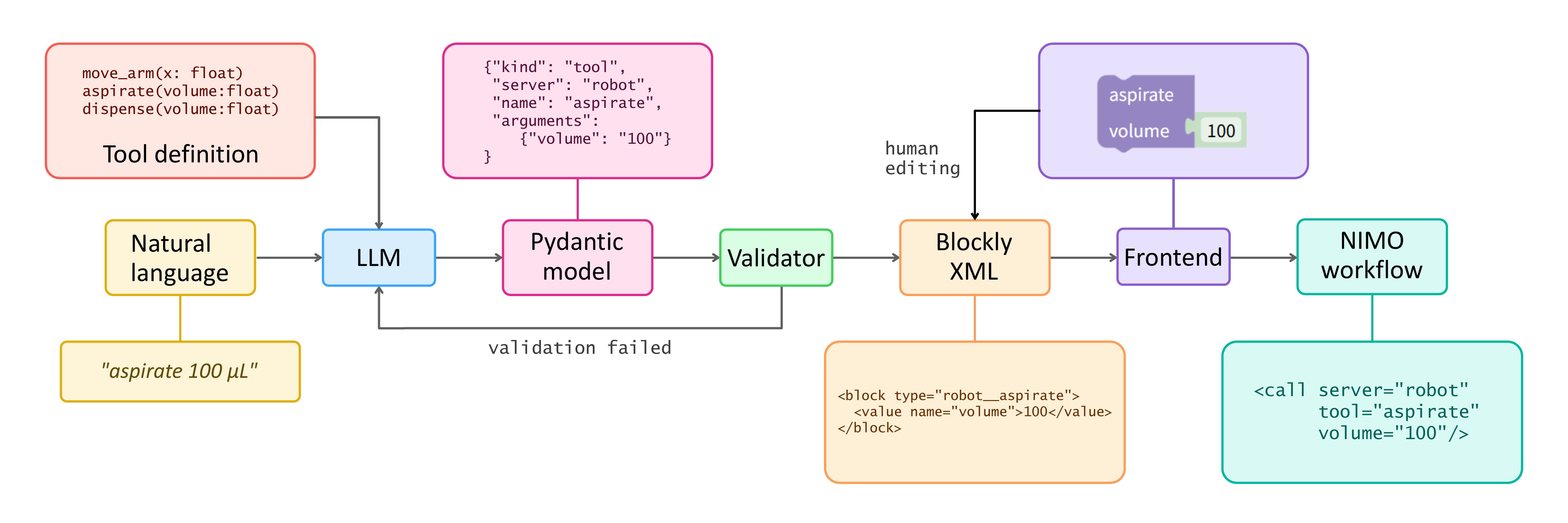}
    \caption{Schematic drawing of the NIMO workflow generation algorithm.}
\label{fig:workflow_generation}
\end{figure}

We use Pydantic AI to ensure that the agent produces structured output.
The agentic workflow generation and execution process is as follows (Figure~\ref{fig:workflow_generation}):

\begin{enumerate}
    \item Based on the user's natural language request, the agent decides whether to invoke the \verb|plan_workflow()| tool, which delegates workflow construction to a dedicated planner LLM.

    \item The planner LLM receives the natural language instruction from the agent and generates structured output conforming to a Pydantic model, based on the available tool definitions and parameter names obtained from the MCP servers.

    \item The validator checks the workflow model against the available MCP servers, tools, arguments, candidate parameters, and loop counters. If validation fails, the error is returned to the planner LLM, which revises the workflow.

    \item After successful validation, a rule-based converter transforms the workflow model into Blockly XML. The frontend loads the XML and displays the corresponding blocks in the visual programming interface.

    \item The user can inspect and edit the displayed blocks. These modifications update the Blockly workspace and its XML representation. At execution time, the frontend exports the current workspace as a workflow abstract syntax tree (AST), which the backend executes without further involvement of the LLM. The workflow executed by the backend is serialized into the NIMO workflow XML format.
\end{enumerate}

This approach builds on the iterative prompting algorithm introduced in our previous work, CLAIRify~\cite{yoshikawa2023large}. However, instead of directly generating XML, the proposed approach uses Pydantic-based structured output and validates the resulting workflow object. We also implemented a direct XML generation approach as a baseline and compared its workflow generation performance with that of the proposed method.

\section{Results}\label{sec:results}
To demonstrate NIMO Controller in a real-world closed-loop setting, we developed a color-matching SDL as a proof-of-concept system. Section~\ref{sec:color_matching} describes its design and illustrates the usage of NIMO Controller.
Section~\ref{sec:user_study} describes the results of a user study using the color-matching SDL.
Section~\ref{sec:workflow} presents an analysis of the workflow generation performance of various LLMs.
\subsection{Color-matching SDL}\label{sec:color_matching}
Color matching has been used as an introductory problem in previous work on orchestration software~\cite{roch2020chemos} and in educational settings~\cite{baird2022minimal}, making it a suitable testbed for evaluating NIMO Controller in a representative closed-loop scenario.

\subsubsection{Hardware setup}
\begin{figure}[H]
\centering
\includegraphics[width=0.5\linewidth]{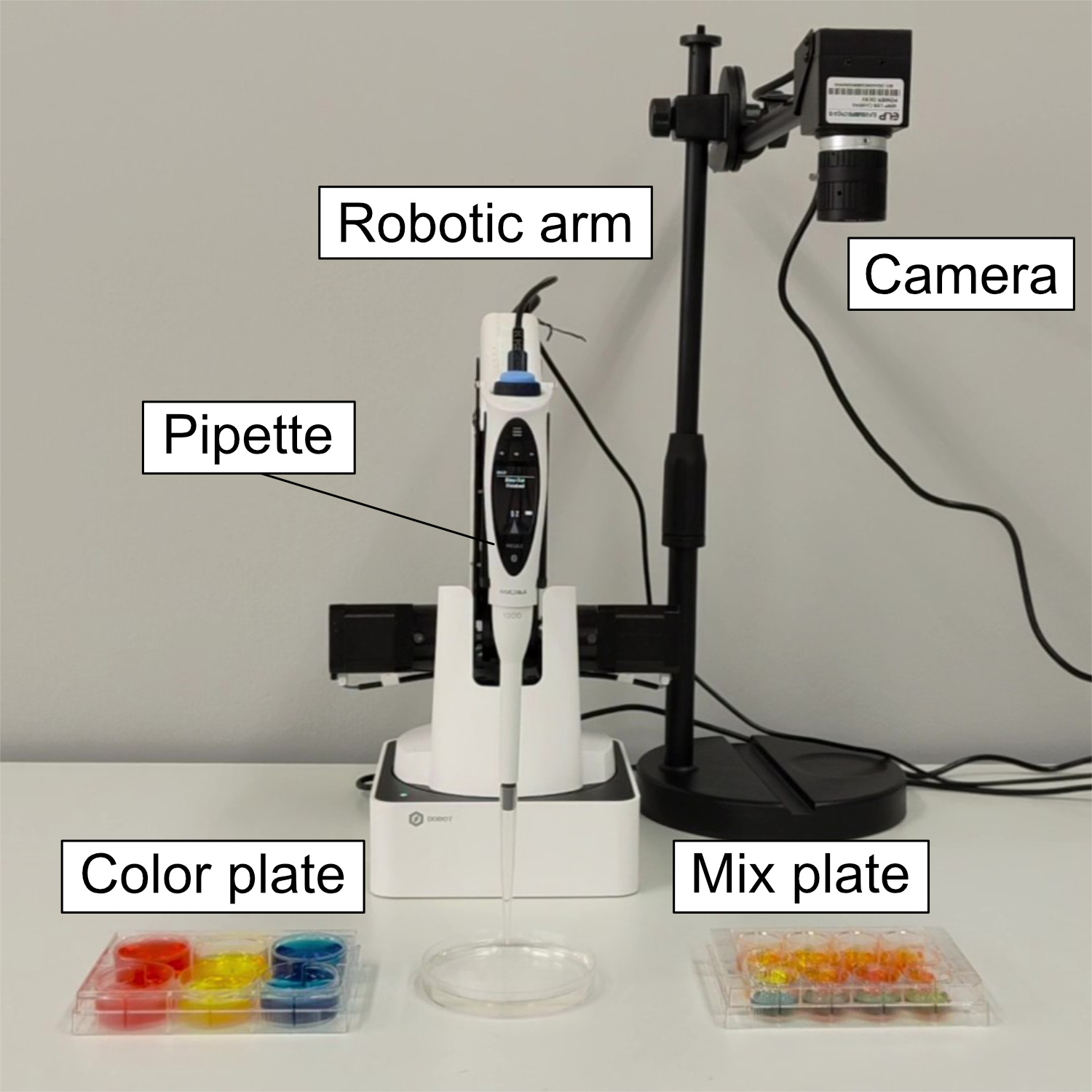}
\caption{Hardware setup of the color-matching SDL.}
\label{fig:setup_color}
\end{figure}

The hardware of this SDL consists of a robotic arm (DOBOT Magician) equipped with an electronic pipette (Picus 2, Sartorius) for color preparation and a USB camera for color measurement (Figure~\ref{fig:setup_color}).
The pipette is mounted on the robotic arm using the jig developed by Hashimoto et al.~\cite{hashimoto2026quantitative}, and both the robotic arm and the pipette are controlled via USB.
Three food coloring dyes (McCormick) corresponding to the primary colors (red, yellow, and blue) are prepared in a 6-well plate with two wells allocated per color, while a separate 12-well plate is used as the mixing workspace.
A Petri dish filled with water is used for washing the pipette tip between dispensing steps.
All components of the experimental setup are placed at fixed locations.
An MCP server exposing functions for pipette control, robotic arm movement, and camera-based color measurement is connected to NIMO Controller.

\subsubsection{Experimental procedure}
The objective of this experiment is to find the mixture ratios of the three dyes that minimize the color difference between the mixed solution and a user-specified target color. The search space consists of dye ratios discretized in 5\% increments, with the three ratios constrained to sum to 100\%. The total dispensed volume is fixed at 2.0 mL. The color of the resulting mixture is captured by the camera as the mean color within a predefined circular region and compared with the target color using the CIEDE2000 color difference formula~\cite{sharma2004ciede}, which provides a perceptually uniform measure of color difference. Users prepare the target color in well 12 at the beginning of the experiment.

The experiment proceeds in a closed-loop fashion using a 12-well plate. Wells 1--4 are used for random exploration to establish an initial dataset, while wells 5--11 are used for optimization guided by PHYSBO~\cite{motoyama2022bayesian, motoyama2026update}, a Bayesian optimization algorithm available in NIMO.

The workflow consists of the following steps: (1) NIMO proposes the dye ratios for the next experiment; (2) the robotic arm dispenses and mixes the dyes in the specified amounts; (3) the camera captures the resulting color; (4) the color difference between the mixture and the target is computed; and (5) the result is fed back to NIMO to update its experiment history. Each experimental iteration takes approximately 12 minutes, dominated by robotic liquid handling.
Users can design the experimental workflow described above through the NIMO Controller visual programming interface, as shown in Figure~\ref{fig:screenshot}.

\begin{figure}[H]
\centering
\begin{subfigure}[t]{0.45\textwidth}
  \centering
  \includegraphics[height=5cm]{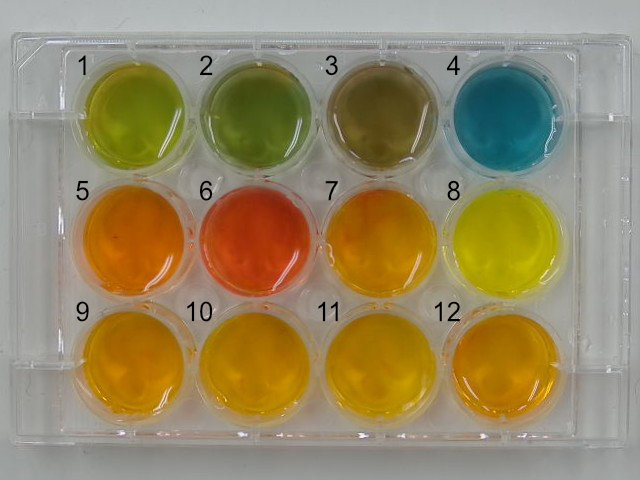}
  \caption{Photograph of the mix plate at the end of the experiment.}
  \label{fig:plate}
\end{subfigure}
\hfill
\begin{subfigure}[t]{0.45\textwidth}
  \centering
  \includegraphics[height=5.5cm]{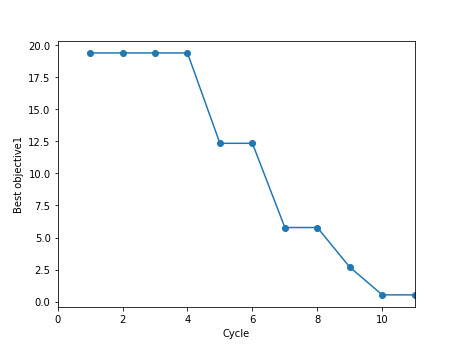}
  \caption{Best color difference at each cycle.}
  \label{fig:graph}
\end{subfigure}

\caption{Results of a color-matching SDL experiment. (a) Photograph of the 12-well plate after the experiment (wells 1--4: random exploration; wells 5--11: Bayesian optimization; well 12: target color). (b) Best CIEDE2000 color difference achieved at each iteration, where lower values indicate a closer perceptual match to the target color.}
\label{fig:color_results}
\end{figure}

Figure~\ref{fig:color_results} shows an example result of a color-matching experiment. The optimization campaign progressively reduced the perceptual color difference over successive iterations while autonomously exploring the three-dimensional discrete parameter space. These results demonstrate that NIMO Controller successfully orchestrates perception, decision-making, and hardware control as a multi-step closed-loop SDL workflow through a visual programming interface automatically generated from MCP servers.

\subsection{User study}\label{sec:user_study}
At an educational event held at a science museum in Tokyo, Japan, we conducted a user study to evaluate the usability of the interface.
We asked 21 junior high and high school students (grades 7--12) to interact with NIMO Controller using the color-matching SDL.
After a 20-minute demonstration and interaction session, we asked the students to complete a questionnaire consisting of the 10-item System Usability Scale (SUS)~\cite{brooke1996sus}, using the Japanese version developed in previous work~\cite{amari2025development}.
Further details of this educational demonstration are reported elsewhere~\cite{yoshikawa2026self}.

\begin{figure}[ht]
\centering
\includegraphics[width=0.95\linewidth]{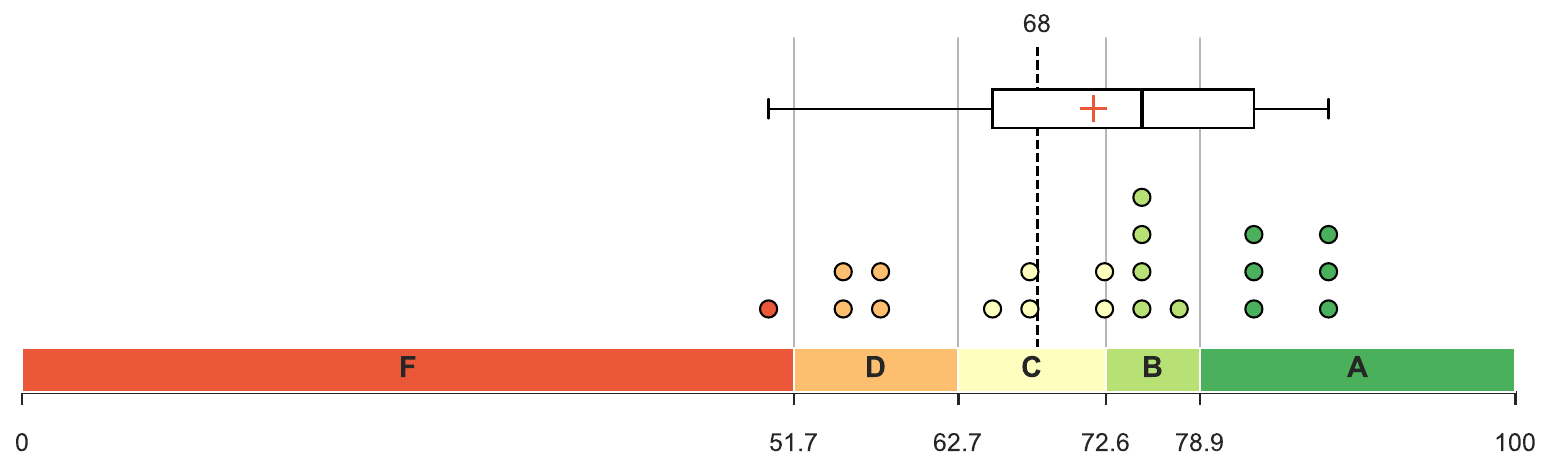}
    \caption{SUS score distribution. Circles denote individual respondents; the boxplot shows the median (black line) and interquartile range, and the red cross indicates the mean value (71.8). The color strip marks the A–F grade bands of the curved grading scale, and the dashed line indicates the average SUS score of 68~\cite{sauro2016quantifying}.}
\label{fig:sus_scores}
\end{figure}

The distribution of SUS scores is shown in Figure~\ref{fig:sus_scores}.
SUS scores range from 0 to 100, and a score of 68 corresponds to the average SUS score across a large number of studies~\cite{sauro2016quantifying}.
The mean SUS score in this study was 71.8 (SD = 11.6), indicating that the developed interface achieved slightly above-average usability.
However, this result should be interpreted with caution because SUS scores may be influenced by cultural and linguistic factors, as discussed in the study on the Japanese version of the SUS~\cite{amari2025development}.

The mean responses to each SUS item are shown in Figure~\ref{fig:sus_per_question}.
Items Q4 and Q10 received the least favorable scores, suggesting that some students felt they needed technical support and had to learn a lot before using the system.
In addition to the inherent complexity of the software, their unfamiliarity with closed-loop experimentation may also have contributed to the less favorable scores for these items.

\subsection{Agentic workflow generation}\label{sec:workflow}
We examined the ability of LLMs to generate NIMO workflows.
We prepared 10 tasks representing typical NIMO workflows, covering different NIMO algorithms and varying levels of workflow complexity.
For each task, we prepared two types of natural language workflow descriptions in English: a \textit{detailed} description that specifies the workflow step by step and a \textit{succinct} description that minimizes the number of words while avoiding ambiguity.
See Appendix~\ref{sec:task_example} for example descriptions.
We evaluated two cloud LLMs (claude-opus-5 and gpt-5.6-luna) and six local LLMs available through Ollama.
The local LLMs were evaluated on a Dell Pro Max with GB10, which is built on the same platform as NVIDIA DGX Spark.
All Ollama settings, including the temperature, were left at their default values.
Workflow generation was considered successful if the sequence of MCP tool calls and their parameters in the generated workflow matched those of the expected workflow.
We set a timeout of 120~s for each task.
As a baseline, we implemented a method similar to CLAIRify~\cite{yoshikawa2023large}, in which the NIMO workflow XML is generated directly.

\begin{figure}[ht]
\centering
\includegraphics[width=0.85\linewidth]{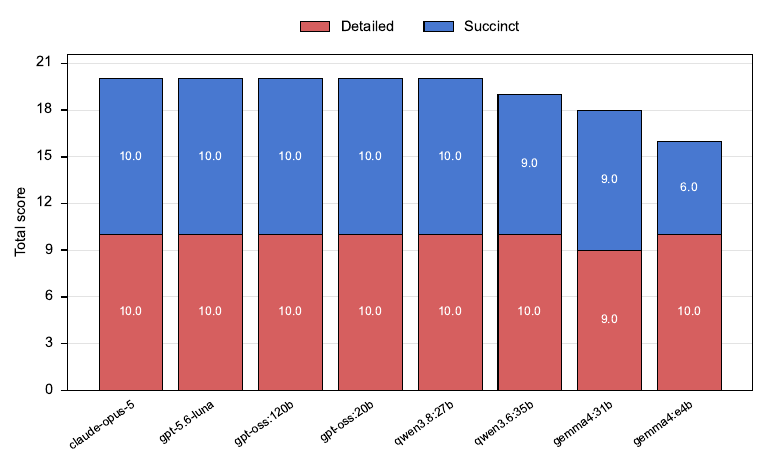}
\caption{Workflow generation performance for different LLMs using Pydantic-based validation.}
\label{fig:llm_evaluation}
\end{figure}

Figure~\ref{fig:llm_evaluation} shows the results of NIMO workflow generation using the proposed Pydantic-based approach.
Notably, cloud models and some local models achieved a 100\% success rate across all tasks, demonstrating that the proposed approach can enable reliable workflow generation even with locally deployed LLMs.
In contrast, other local models showed lower performance, indicating that workflow generation performance still depends on the capability of the underlying LLM.
The level of detail in the workflow description had only a minor effect on performance.

\begin{figure}[H]
\centering
\includegraphics[width=0.85\linewidth]{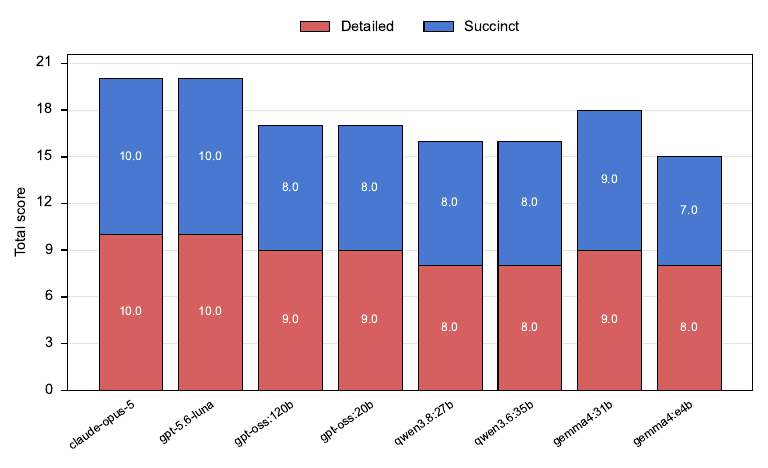}
\caption{Workflow generation performance for different LLMs with direct XML generation approach, which is similar to CLAIRify~\cite{yoshikawa2023large}.}
\label{fig:llm_evaluation_clairify}
\end{figure}

For comparison, Figure~\ref{fig:llm_evaluation_clairify} shows the results obtained using the direct XML generation approach.
Although both cloud LLMs still achieved a 100\% success rate, none of the local LLMs achieved perfect performance, and they generally performed worse than with the Pydantic-based approach.
These results suggest that the Pydantic-based approach is more robust, particularly when using local LLMs.


\section{Conclusion and future perspectives}\label{sec:conclusion}
In this paper, we have proposed an MCP-based SDL software architecture and introduced NIMO Controller as an orchestrator for such SDLs. It enables no-code algorithmic decision-making and automated experimentation through both visual programming and natural language interfaces.
We demonstrated the system in a color-matching SDL case study, in which a complete closed-loop workflow was deployed. The SUS results indicated acceptable usability in an educational setting. We also evaluated agentic workflow generation across different LLMs and found that highly reliable workflow generation can be achieved with sufficiently capable LLMs.

NIMO Controller adopts Blockly to facilitate the implementation of the visual programming interface and to provide an interface familiar to students with experience using Scratch. However, this design also limits the expressiveness of NIMO workflow representations. In particular, Blockly is not well suited to representing complex data flows and workflow structures. One significant limitation is the handling of multiple return values: the current Blockly-based interface cannot readily represent multiple distinct outputs from a single MCP tool call, making it difficult to use each output independently in subsequent workflow steps. Furthermore, the block-based representation becomes increasingly cumbersome as workflows grow in length and complexity.

Future work includes developing a more expressive workflow representation capable of handling more complex data flows and workflow structures while maintaining the safety and controllability of the current MCP-based approach. We also plan to develop an agent that can generate workflows using this representation and visualize them in an interactive interface, allowing users to inspect and modify agent-generated workflows before execution.

In addition, the current implementation is primarily designed to leverage the existing functionality of NIMO, particularly for experimental design and result visualization. However, LLMs themselves exhibit emergent capabilities in experimental planning, as explored in our previous work~\cite{tamura2026llm}, and their coding capabilities can provide more task-specific and flexible visualization of experimental results. Future development of NIMO Controller will therefore focus on integrating these capabilities into the workflow framework, combining the reliability and controllability of algorithmic experimental planning with the flexibility of more autonomous LLM-based agents while retaining explicit workflow representation and user oversight.

In this work, MCP tools for the hardware devices were implemented using the Python interfaces provided by their manufacturers. A promising direction for future work is to integrate MCP with device-integration standards such as LADS OPC UA and SiLA 2, as well as the recently introduced Model Hardware Standard (MHS). These standards can provide fine-grained device control, while MCP offers a higher-level interface for orchestration and AI-agent integration. Combining these complementary layers could further expand the interoperability and applicability of MCP-based orchestration.

\section*{Software availability}
NIMO Controller is available on GitHub under the MIT license at \url{https://github.com/NIMS-DA/nimo-controller}.

\section*{Acknowledgments}
This work was supported by JSPS KAKENHI Grant Number JP25K21333 and JST PRESTO Grant Number JPMJPR24T8.

\bibliography{reference}
\bibliographystyle{unsrtnat}
\newpage
\appendix
\setcounter{figure}{0}
\renewcommand{\thefigure}{\thesection\arabic{figure}}
\section{SUS results by question}~\label{sec:sus_questions}
\begin{figure}[htbp]
\centering
\includegraphics[width=1.0\linewidth]{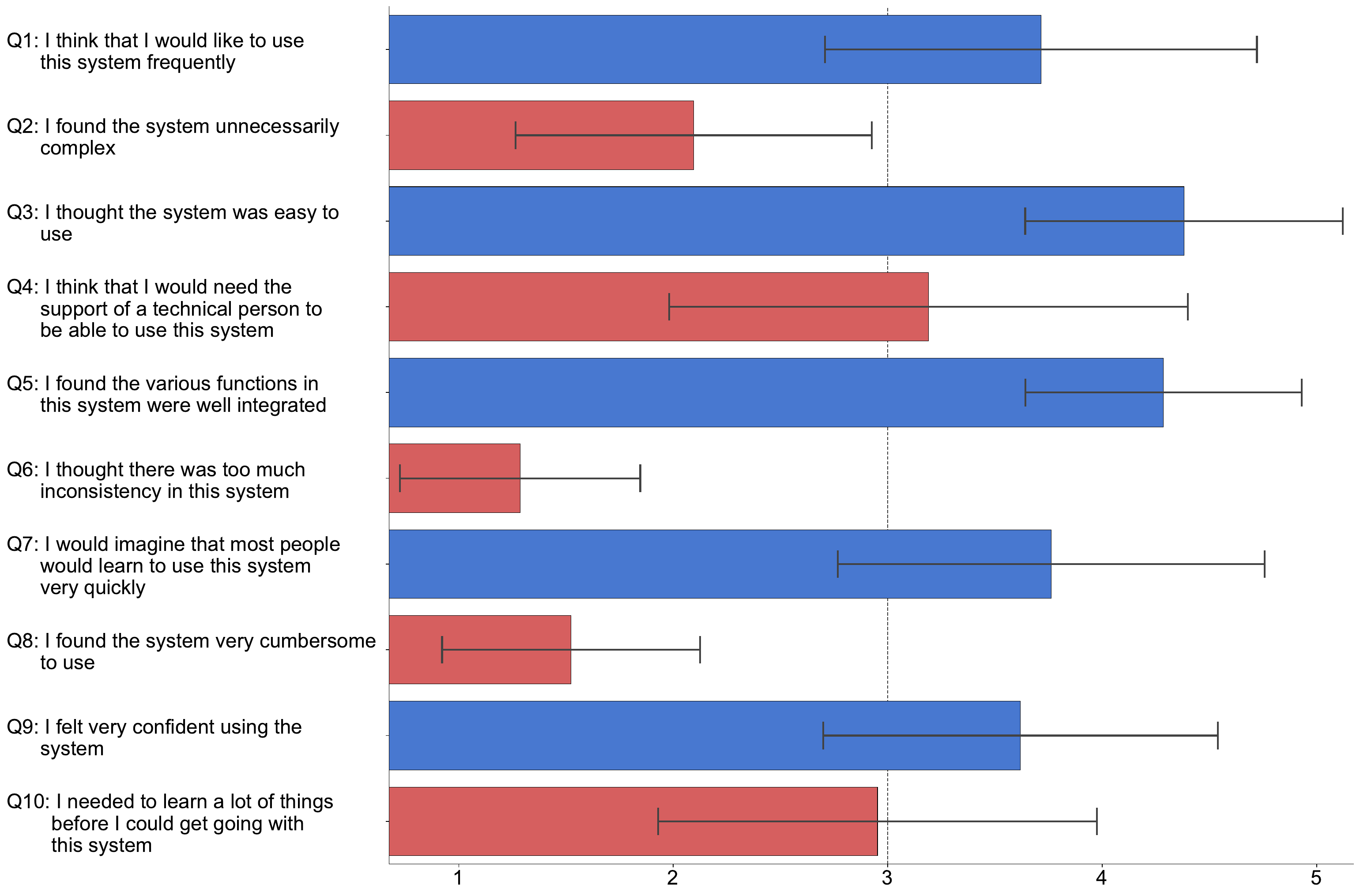}
    \caption{Mean responses to the ten SUS items as recorded on the original 1–5 scale. For odd-numbered items (blue), higher scores are better, while lower scores are better for even-numbered items (red). Error bars show $\pm$1 SD.}
\label{fig:sus_per_question}
\end{figure}
\newpage

\section{Workflow generation task examples}~\label{sec:task_example}
\setcounter{lstlisting}{0}
\renewcommand{\thelstlisting}{\thesection\arabic{lstlisting}}
\begin{lstlisting}[
  caption={Example of a detailed description of a NIMO workflow for the color-matching experiment (Listing~\ref{lst:nimo-workflow}).},
  float=ht
]
We will run a color-matching experiment on the mixing robot, minimizing the color difference to the target color 00ffff. First, move the robot to the home position once. Then perform exactly 12 mixing cycles. In each of the first 4 cycles, choose the next candidate using the selection tool with the RE (random exploration) method. In each of the remaining 8 cycles, choose the next candidate using the minimization tool with the PHYSBO method. In every cycle, handle the three colors one at a time, in the order red, blue, yellow. For each color: move to that color's well, aspirate the proposed volume of that color (proposal.red, proposal.blue or proposal.yellow), move to the mixing well whose number equals the current cycle counter (starting from 0), and dispense the same proposed volume there. After all three colors have been dispensed, evaluate get_color_diff on the current cycle's mixing well against the target hex color 00ffff and register the returned value as the objective via an update. After all 12 cycles are finished, plot the history of the best objective value in minimization mode.
\end{lstlisting}

\begin{lstlisting}[
  caption={Example of a succinct description of a NIMO workflow for the color-matching experiment (Listing~\ref{lst:nimo-workflow}).},
  float=ht
]
Match the target color 00ffff in 12 cycles by minimizing get_color_diff. First, move to the home position. The first 4 cycles use RE selection, the remaining 8 use PHYSBO minimization. In each cycle, for red, blue, yellow in that order: move to the color well, aspirate the proposed volume, move to the mixing well numbered by the cycle counter, and dispense it. Then evaluate get_color_diff on that well against 00ffff and update the objective. Finally, plot the best-value history in minimization mode.
\end{lstlisting}
\end{document}